# Efficient Training of Spiking Neural Networks with Temporally-Truncated Local Backpropagation through Time

Wenzhe Guo, *Student Member, IEEE*, Mohammed E. Fouda, *Member, IEEE,* Ahmed M. Eltawil, *Senior Member, IEEE,* and Khaled Nabil Salama, *Senior Member, IEEE*

**Abstract—Directly training spiking neural networks (SNNs) has remained challenging due to complex neural dynamics and intrinsic non-differentiability in firing functions. The well-known backpropagation through time (BPTT) algorithm proposed to train SNNs suffers from large memory footprint and prohibits backward and update unlocking, making it impossible to exploit the potential of locally-supervised training methods. This work proposes an efficient and direct training algorithm for SNNs that integrates a locally-supervised training method with a temporally-truncated BPTT algorithm. The proposed algorithm explores both temporal and spatial locality in BPTT and contributes to significant reduction in computational cost including GPU memory utilization, main memory access and arithmetic operations. We thoroughly explore the design space concerning temporal truncation length and local training block size and benchmark their impact on classification accuracy of different networks running different types of tasks. The results reveal that temporal truncation has a negative effect on the accuracy of classifying frame-based datasets, but leads to improvement in accuracy on dynamic-vision-sensor (DVS) recorded datasets. In spite of resulting information loss, local training is capable of alleviating overfitting. The combined effect of temporal truncation and local training can lead to the slowdown of accuracy drop and even improvement in accuracy. In addition, training deep SNNs' models such as AlexNet classifying CIFAR10-DVS dataset leads to 7.26% increase in accuracy, 89.94% reduction in GPU memory, 10.79% reduction in memory access, and 99.64% reduction in MAC operations compared to the standard end-to-end BPTT.**

## I. INTRODUCTION

IN recent years, deep learning surged as a method for solving various complex tasks, such as visual processing[1], language processing [2], object detection[3], and medical diagnostics [4], making it the most promising and dominant approach. The remarkable performance of deep learning comes at the expense of substantial energy consumption resulting from intensive full-precision matrix multiply-accumulate (MAC) operations in artificial neural networks (ANNs). This drawback holds back deep learning algorithms from being deployed on resource-constrained platforms, such as edge devices. Inspired by the biological nervous system, spiking neural networks (SNNs) have attracted ever-growing attention from research communities for their superior energy efficiency to ANNs. Information in SNNs is transmitted and processed on the occurrence of a spike or an event. The large spike sparsity and simple synaptic operations in SNNs give rise to low energy consumption. SNNs have been explored in a broad range of applications, such as pattern recognition[5, 6], object detection [7], navigation [8], and motor control [9]. Based on SNNs, neuromorphic computing systems have been proposed as an alternative computing paradigm to the traditional Von Neumann systems [10-12].

Training SNNs has been a significant challenge in exploiting the full potential of SNNs due to complex neural dynamics and discontinuous spikes [13]. The lack of efficient and effective training algorithm limits the use of SNNs in complex real-world tasks. Existing training algorithms can be categorized into two general approaches: indirect training and direct training. The indirect training relies on the conversion from a well-trained ANN model to an architecturally equivalent SNN model. The learned parameters in the DNN are directly transferred to the SNN, while the activations in the DNN corresponds to the firing rates of SNN neurons [14-16]. Direct training methods can be categorized into unsupervised and supervised approaches. The unsupervised training methods, such as spike-timing-dependent plasticity (STDP), are inspired by the biological nervous systems, modifying weights in terms of local synaptic activities [17]. Without supervision signals, these methods exhibit inferior performance [18, 19]. The supervised training methods are mainly based on gradient descent optimization, such as SpikeProp [20] and Tempotron [21]. Another type of supervised method derives a transfer function that formularizes the accumulated effect of spikes, like firing activity or rate, from the event-based update of membrane potential [22-24]. Due to the similarity between SNNs and recurrent neural networks (RNNs), it is not surprising that the training algorithm, backpropagation through time (BPTT), used in RNNs can be borrowed for SNNs [25]. The training process is depicted in **Fig. 1** (a). During the forward process, neural

This work was funded by the King Abdullah University of Science and Technology (KAUST) AI Initiative, Saudi Arabia. (Corresponding author: Khaled Nabil Salama.)

Wenzhe Guo, Ahmed M. Eltawil, and Khaled Nabil Salama are with the Division of Computer, Electrical and Mathematical Sciences and Engineering, King Abdullah University of Science and Technology, Thuwal 23955, Saudi Arabia (e-mail: wenzhe.guo@kaust.edu.sa; ahmed.eltawil@kaust.edu.sa; khaled.salama@kaust.edu.sa).

Mohammed E. Fouda is with CECS, University of California–Irvine, Irvine, CA 92612 USA (e-mail: foudam@uci.edu).

states in SNNs are iteratively updated with both spatial and temporal inputs throughout the whole time window. The backward process starts at the end of the training window when the loss function is computed. BPTT has been demonstrated to be very effective in training SNNs by considering the spatio-temporal dynamics [26-31]. SNNs trained with BPTT closely approach ANNs in classification performance on various datasets. More importantly, BPTT allows SNNs to be scaled to very deep networks (50 layers) and hence empowers SNNs to solve more complex tasks.

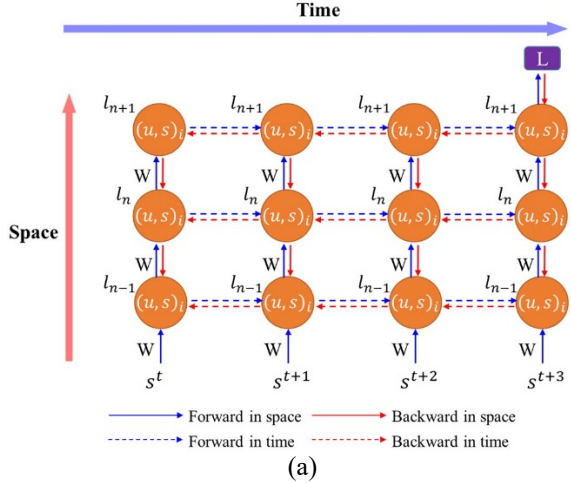

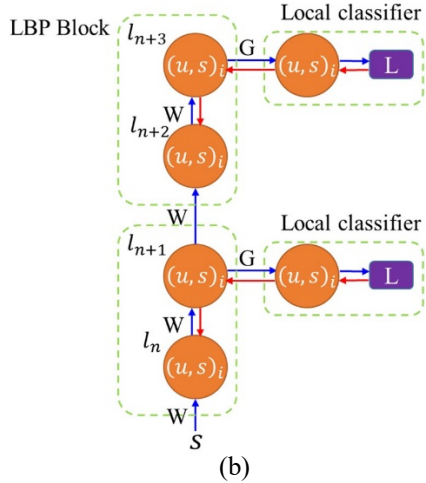

**Fig. 1** (a) Training SNNs with backpropagation through time. Training processes are unfolded in time. (b) Training SNNs with local backpropagation. Each node represents a spiking neuron layer where $l$ is the layer index, $i$ is the neuron index, $u$ is the membrane potential, and $s$ is the spike. $W$ is weights between layers in the main network and $G$ is the weights between a network layer and a local classifier. $L$ is a loss function.

Typically, in the standard BP algorithm, errors are propagated backward in a layer-by-layer fashion to update training parameters. The activation values need to be saved during the forward pass and read out for parameter updates during the backward pass. Despite the effectiveness of the standard BP algorithm, the network suffers from frequent memory access, computational inefficiency, and long training time. Thus, various local BP (LBP) algorithms are proposed to tackle the aforementioned issues [32-35]. An example of block-wise LBP training is illustrated in **Fig. 1** (b). LBP algorithms attach a classifier to a block of layers (or a single layer) in the network and train each block separately and simultaneously. Since training happens locally, intermediate states can be saved in buffers temporally before parameter updates, eliminating the need for memory storage and access [32]. Moreover, LBP divides the whole network into gradient-isolated modules, making the hardware design scalable because the network can be built by cascading the same local training module. However, LBP suffers from inferior performance compared to the standard BP because of information loss [35]. Few works have considered LBP in SNNs. Neftci *et al.* demonstrated the effectiveness of LBP in SNNs [36]. In his work, a classifier with random weights was attached to each layer. The networks were trained by approximate BP with a surrogate gradient for the firing function at each time step. Temporal dependency in the backward update was completely ignored because of intractable gradient computation in the algorithm. Although competitive classification performance was achieved on one dataset against the state-of-the-art works, the method required to learn each input image in a very long time window, 500 time steps. This work failed to provide a fair comparison against the BPTT algorithm and generalize the effectiveness of the proposed training algorithm to deeper networks and complex datasets. Ma *et al.* experimented layerwise local training in spike-based BPTT to train SNNs [37]. Good performance was achieved in various tasks but at the expense of high computational cost.

The BPTT algorithm dictates that the backward pass can only happen after the network moves forward throughout the whole time window. It requires the network to store the time evolutions of neural states as the backward pass needs them to compute gradients at each step, which incurs a substantial memory footprint. The accumulation of gradients in a long time window can cause gradient exploding issues [38]. When the local training method is applied together with BPTT in SNNs, the benefits of LBP are largely diminished due to the fact that BPTT prohibits the parallel execution of forward pass and backward pass. Except for the last time step, parallel execution in LBP is impossible and intermediate states have to be stored in external memory. As a result, LBP loses its advantage over standard BP. Inspired from the idea in truncated BPTT (TBPTT) applied in RNNs [39, 40], we introduce temporal truncation in spiking BPTT to resolve the incompatibility issue between LBP and BPTT for training SNNs. TBPTT divides the training time window into many temporal chunks and runs BPTT for each chunk. It breaks the temporal restriction imposed on the backward pass, allowing for the advantages of LBP to be considerable. The smaller the chunk, the more significant contribution LBP can make. Additionally, temporal truncation is able to cut short computational graphs for backward updates proportionally, leading to significant reduction in memory footprint.

In this work, we propose an efficient training method for SNNs by integrating local training methods with BPTT by introducing temporal truncation. The proposed method can significantly reduce memory footprint and access, and arithmetic operations with negligible performance loss. The training process can benefit from the proposed method both temporally and spatially. However, both LBP and TBPTT could suffer from inferior performance depending on the size of truncated chunks and the length of local blocks. Thus, in this work, we will investigate the impact of temporal truncation and BPTT on classification performance and computational cost in SNNs.

The main contributions are summarized as follows.

1) We introduce temporal truncation in BPTT to resolve the incompatibility issue between LBP and BPTT, and thus propose an efficient training algorithm for SNNs with significantly reduced memory footprint and access, and arithmetic operations.

2) We thoroughly explore the design space regarding the temporal truncation length and local training size and analyze their impact on classification performance and computational cost of different SNNs for various datasets.

3) We provide analytical models for predicting and estimating memory footprint and access, arithmetic operations on different hardware platforms.

4) We compare trainable classifiers and random classifiers applied in LBP and demonstrate that random classifiers do not provide considerable advantages while suffering from severe performance drop.

The rest of this article is organized as follows. Section II introduces neural models and the proposed training algorithm. Section III describes the details of experiments and presents classification results. Section IV analyzes computational cost of the proposed algorithm and presents corresponding results. Section V summarizes our work and discusses limitations and future perspectives.

## II. METHODS

### A. Neural models

Leaky integrate-and-fire neuron (LIF) model is widely used to model spiking neurons because it can accurately capture neural dynamics and has excellent computational efficiency. It consists of a linear equation and a threshold condition. The model can be written as

$$u_i^{t+1,n} = \tau u_i^{t,n} + \sum_j W_{ij}^n s_j^{t+1,n-1} - \theta s_j^{t,n} \qquad (1)$$

$$s_i^{t+1,n} = \Theta(u_i^{t+1,n} - u_{th}) \qquad (2)$$

where $u_i^{t,n}$ and $s_i^{t,n}$ are the membrane potential and output spike of the neuron $i$ in the layer $n$ at time $t$, respectively, $W_{ij}^n$ is the synaptic weight between the neuron $i$ in the layer $n$ and the neuron $j$ in the layer $n-1$, $\tau$ is the leaky time constant, $u_{th}$ is the threshold potential, $\theta$ is the reset constant, and $\Theta(\cdot)$ is a unit step function. A soft reset mechanism is used to reset membrane potential once an output spike is generated.

### B. Temporally-truncated local BPTT

We introduce temporal truncation in the BPTT algorithm together with local classifiers to jointly train SNNs. **Fig. 2** illustrates the proposed training method, where temporal truncation with a step size of 2 and two-layer local blocks are applied in BPTT. During forward pass, neuron states of the main network and local classifiers are updated iteratively in space and time, as indicated by blue arrows. Backward pass happens after every truncation interval. A loss is computed at each local classifier during the backward pass, and errors are propagated backward from classifiers spatially to local blocks and temporally to their previous states, as indicated by the red arrows. Inside a local block, errors are propagated in the same fashion. But the error flow stops between blocks, removing the backward dependency between blocks. This way eliminates the need to store intermediate states of the current block in external memory and makes it possible to execute forward pass and backward pass in parallel. The error flow is also discontinued between truncation intervals, which eliminates the need to store all the neural states updated in the previous intervals. The training process benefits from both temporal truncation and local learning in reducing the computational cost. However, both methods could also affect network performance, since temporal truncation removes the temporal dependency between truncated intervals during the backward pass and local training could potentially cause information loss. Accordingly, we define a variable pair $(k, n)$, as the length of a temporal interval and the number of layers in a local block, respectively. We explore the design space of these two factors and analyze the impact on network performance.

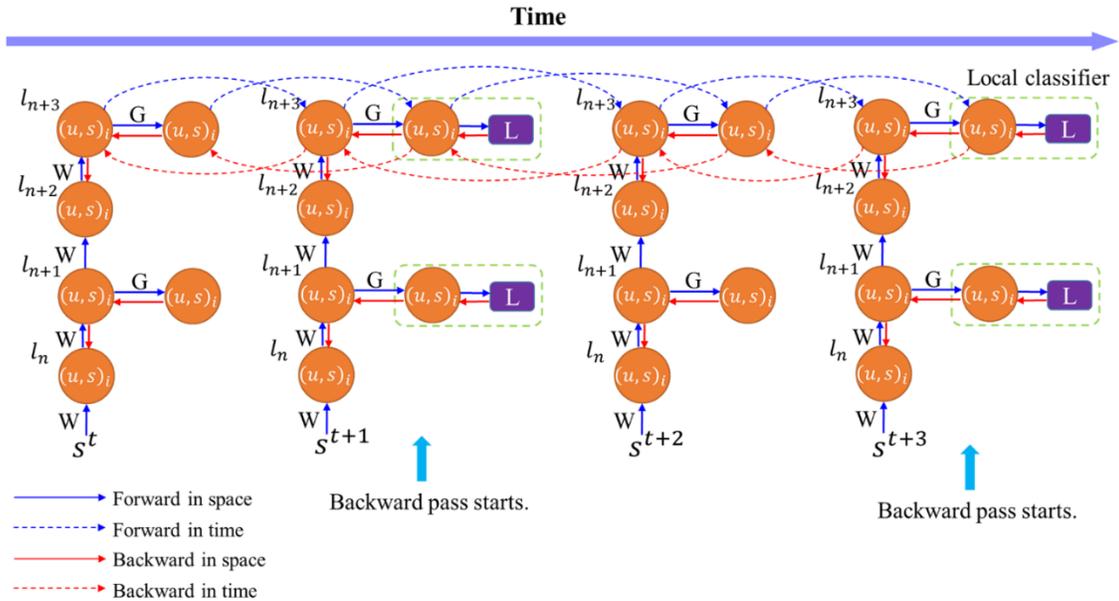

**Fig. 2** Training process of an SNN with temporally-truncated local BPTT unfolded in time. In this illustration, the length of the truncation interval and local block are both 2. The forward and backward update flows are indicated by the blue and red arrows, respectively. The update flows in time (dashed arrows) are only depicted for the top layer for clarity. $W$ represents the weights between layers in the main network, and $G$ is the weights between local blocks and classifiers.

TABLE 1
HYPERPARAMETER SETTING

| Hyperparameter | EMNIST | DvsGesture | CIFAR10 | CIFAR10-DVS |
|---|---|---|---|---|
| Batch size | 1024 | 32 | 128 | 128 |
| Momentum | 0.9 | 0.9 | 0.9 | 0.9 |
| Time steps, $T$ | 20 | 60 | 10 | 100 |
| Gradient width, $a$ | 0.5 | 0.5 | 0.5 | 0.5 |
| Dropout rate | 0 | 0-0.2 | 0-0.15 | 0-0.2 |
| Learning rate | 0.2-0.5 | 0.005-0.2 | 0.1-0.2 | 0.1-0.5 |
| Leaky constant, $\tau$ | 0.9 | 0.3 | 0.8 | 0.8 |
| Threshold potential, $u_{th}$ | 0.4-0.6 | 0.2-0.5 | 0.3-0.5 | 0.3-0.6 |

Moreover, in local training methods, applying trainable local classifiers can retain high performance but add additional weight parameters to be trained, incurring training overhead. Using random weights in local classifiers was proposed to reduce the overhead but proven to be less effective in training networks [32]. Thus, we provide a detailed analysis of the effect of trainable and random local classifiers in our proposed training algorithm.

Following the theoretic framework in [26], we derive the essential equations used in the training algorithm as follows. Firstly, we define the loss function as the mean squared error between the time average firing rates of classifier neurons and target firing rates, expressed by

$$L = \frac{1}{N_c} \sum_i \left( y_i - \frac{1}{k} \sum_t s_{c,i}(t) \right)^2 \qquad (3)$$

where $y_i$ and $s_{c,i}(t)$ is the target firing rate of classifier neuron $i$ and the actual firing rate at time $t$, respectively, and $N_c$ is the number of classes. The average firing rate is calculated over each truncation interval. In this work, the target firing rate vector for the classifier is determined as a one-hot vector based on the target class. Then, we define the spike error, $\delta_i^{t,n} = \partial L / \partial s_i^{t,n}$, and the potential error, $\gamma_i^{t,n} = \partial L / \partial u_i^{t,n}$. Based on the two errors, the iterative backward update equations are given as,

$$\delta_i^{t,n} = \sum_m \gamma_m^{t,n+1} W_{mi}^n + \gamma_i^{t+1,n}(-\theta) \qquad (4)$$

$$\gamma_i^{t,n} = \delta_i^{t,n} \Theta'(u_i^{t,n} - u_{th}) + \gamma_i^{t+1,n} \tau \qquad (5)$$

where the first component on the right side of the equations is contributed by the errors propagated spatially from the upper layer, and the second component is due to the temporal error backpropagation. Clearly, the potential error needs to be propagated backward in space and time. Different surrogate gradient functions were proposed to solve the discontinuity issue with the firing function in (2) [26]. A rectangular function is proven to be effective and simple, and thus the gradient function can be approximated by

$$\Theta'(u) \approx \frac{1}{a} sign\left( |u - u_{th}| < \frac{a}{2} \right) \qquad (6)$$

where $sign(\cdot)$ is the sign function, and $a$ is the width of the non-zero region. In TBPTT, weight gradients are accumulated over the truncation interval by summing up all the gradients computed at each time step, expressed as

$$\frac{\partial L}{\partial w_{ij}^n} = \sum_t \gamma_i^{t,n} s_j^{t,n-1} \qquad (7)$$

where the summation goes over the truncation interval. With the computed gradients, weight parameters can be updated by an optimization method, such as stochastic gradient descent (SGD) and Adam [41, 42]. The implementation details of the training algorithm are explained in Algorithm 1.

## III. Experiments

### A. Experimental setup

We evaluate the proposed training algorithm on two different spiking convolutional neural networks (SCNNs) adapted from LeNet and AlexNet architectures [43, 44]. Each network was tested on two different types of datasets: static frame-based datasets and dynamic-vision-sensor (DVS) recorded datasets. LeNet was used to classify extended MNIST (EMNIST) dataset [45] and DvsGesture dataset [46], while AlexNet was used to classify CIFAR10 dataset [47] and CIFAR10-DVS dataset [48]. Simulations were performed using PyTorch framework [49]. The mean squared error (MSE) loss function and SGD optimization method with momentum were used for updating parameters [42]. As for regularization, a dropout layer was added after each convolutional or fully-connected layer [50]. Accuracy results were recorded after 100 training epochs for all the simulations. A step-decay scheduling method was used to reduce learning rate by a factor of 2 every 20 epochs. The other hyper-parameters used in all the experiments are listed in TABLE 1. We vary the values of $(k, n)$ and obtained corresponding classification accuracy on each dataset.

*B. Spiking convolutional neural networks*

**Algorithm 1** Training algorithm for one batch iteration.

---

**Inputs:** Network inputs $\{X^t\}_{t=1}^T$, class labels Y, parameters and initial neural states of the main network $\{W^l, U^{t_0,l}, S^{t_0,l}\}_{l=1}^{N_L}$, parameters and initial neural states of local classifiers $\{W_c^l, U_c^{t_0,l}, S_c^{t_0,l}\}_{l=1}^{N_C}$, training window $T$, training variables $(k, n)$, other hyper-parameters.

**Initialize** all the parameters and neural states.

for interval $i = 1$ to $T/k$ do

  **Forward pass:**

  for time $\tau = 1$ to $k$ do

    $t = \tau + (i - 1) * k$

    for layer $l = 1$ to $N_L$ do

      Network state update: $\{U^{t,l}, S^{t,l}\} \leftarrow$ Update$\{W^l, U^{t-1,l}, S^{t,l-1}, S^{t-1,l}\}$, from (1), (2).

      if $l \% n == 0$,

        Classifier state update: $\{U_c^{t,l}, S_c^{t,l}\} \leftarrow$ Update$\{W_c^l, U_c^{t-1,l}, S_c^{t,l-1}, S_c^{t-1,l}\}$, from (1), (2).

      end if

    end for

  end for

  for layer $l = 1$ to $N_L$ do

    if $l \% n == 0$,

      Compute loss: $L \leftarrow$ Loss function$\{Y, \sum_t S_c^{t,l}\}$, from (3)

    end if

  end for

  **Backward pass:**

  for time $\tau = k$ to $1$ do

    $t = \tau + (i - 1) * k$

    for layer $l = N_L$ to $1$ do

      if $l \% n == 0$,

        Compute errors and gradients at classifiers.

        Accumulate gradients.

      end if

      Backpropagate errors and compute gradients: from (4), (5), (7)

      Accumulate gradients.

    end for

  end for

  Update weights.

end for

---

The network structures used in the experiments are listed in TABLE 2. The spiking LeNet is a four-layer spiking CNN adapted from the original LeNet, consisting of three convolutional layers, two average-pooling layers, and one fully-connected layer. LIF neuron layer is placed after each of these layers, so that each layer outputs spikes. A classifier is not included in the table. We used two network scales. The smaller network, LeNet-1, with a smaller number of channels and neurons, was tested on EMNIST, while the larger network, LeNet-2, was for DvsGesture because of pattern complexity. We also constructed an eight-layer spiking AlexNet with similar network settings to the original AlexNet. It consists of six convolutional layers followed by two average-pooling layers, and two fully-connected layers.

TABLE 2
Network Architectures

| **LeNet-1/2** | | |
|---|---|---|
| Layer type | # Channels or neurons | (Kernel size, stride) |
| Conv | 6/64 | (5, 1) |
| AvgPool | 6/64 | (2, 2) |
| Conv | 16/128 | (5, 1) |
| AvgPool | 16/128 | (2, 2) |
| Conv | 120/128 | (5, 1) |
| FC | 128/256 | NA |
| **AlexNet** | | |
| Layer type | # Channels or neurons | (Kernel size, stride) |
| Conv | 96 | (3, 2) |
| Conv | 256 | (3, 1) |
| AvgPool | 256 | (2, 2) |
| Conv | 384 | (3, 1) |
| AvgPool | 384 | (2, 2) |
| Conv | 512 | (3, 1) |
| Conv | 384 | (3, 1) |
| Conv | 256 | (3, 1) |
| FC | 4096 | NA |
| FC | 1024 | NA |

*C. Encoding methods*

**Frame-based datasets**. Since the images from both EMNIST and CIFAR10 datasets are comprised of integer-valued pixels, they are incompatible with SNNs. The widely used conversion method is rate encoding, which converts each pixel into a spike train with a frequency proportional to the pixel intensity. This method suffers from high training latency and precision loss. Many works proposed a direct encoding method that uses the first layer as an encoding layer, directly receiving intensity values and outputting spikes as inputs to the next layer [27, 30, 51]. This method largely reduces training latency and retains good performance. Although the first layer computes as an ANN layer, under the fact that networks generally consist of tens or hundreds of layers, this has little impact on the computational efficiency of SNNs. Thus, we adopted the direct encoding method in our experiments.

**DVS-recorded datasets**. DVS camera produces event streams encoded in timestamps, xy coordinates, and polarity. SNNs can not directly process the raw data. We converted each encoded event streams into a time series of event images with two channels and binary pixel intensity. The two channels correspond to the polarity of events, and the binary intensity indicates the occurrence of an event at the pixel location. Due to long recording time and high resolution, we accumulated all the event images in a defined time window $\Delta t$ into one new image and took the first $T$ new images as inputs to SNNs. The values of $(\Delta t, T)$ are $(20\ ms, 60)$ for DvsGesture dataset, and $(5\ ms, 100)$ for CIFAR10-DVS dataset.

### D. Classification accuracy results

Temporal truncation ignores the temporal dependency spanning across truncated intervals in the backward pass, which introduces bias on short-term dependency. The local training method utilizes local errors to learn features that only benefit advantages, these two methods could have a negative impact on classification performance. Thus, it is necessary to investigate and analyze their roles. In Section II-B, we defined a variable pair $(k, n)$, as the length of the truncated temporal interval and the number of layers in a local block, respectively. The case where $k$ equals the total time step $T$ refers to the standard BPTT, whereas $k=1$ suggests no temporal dependency in the backward pass. In this section, we experiment with different sets of $(k, n)$ in each classification task and analyze the change of classification accuracy.

For classification on EMNIST dataset, $k$ was chosen from the set, $\{20, 10, 5, 2, 1\}$, and $n$ from the set, $\{4, 2, 1\}$. For classification on CIFAR10 dataset, $k$ was chosen from the set, $\{10, 5, 2, 1\}$, and $n$ from the set, $\{8, 4, 2, 1\}$. The accuracy results on the two datasets are shown in **Fig. 3** and **Fig. 4**, respectively. Since the spiking LeNet is of four layers, $n = 4$ corresponds to the standard BP. And $n = 8$ in AlexNet also corresponds to the standard BP. We use *LBPn* to indicate the local BP with $n$ layers in each local block. From the results, it can be seen that in most cases, accuracy tends to decrease with the truncation interval, especially when the interval is small. In the classification experiment on EMNIST, local learning methods are affected by temporal truncation more significantly than standard BP. BP shows the best accuracy compared against LBP2 and LBP1, which suggests that LBP causes the loss of useful information. However, in the classification experiment on CIFAR10, BP is more severely affected by temporal truncation with a 6.30% accuracy drop compared with the maximum drop of 3.35% for LBPs, as indicated in **Fig. 4** (a). LBP4 shows the best results in most cases. This can be due to that LBP alleviates the overfitting effect [52]. LBP divides the network into smaller blocks and trains each block separately for the same task. In some way, the actual number of parameters required to learn features for the task is reduced, which leads to less severe overfitting. The combined effect of temporal truncation and local training can be observed in **Fig. 4**, where the accuracy of BP is affected by temporal truncation more severely than LBPs. LBP2 has lower accuracy at large intervals, but gets closer to and even surpasses BP when the interval shrinks. In the case of LBP1, the accuracy is improved all the way. Comparing the results obtained from trainable classifiers and random classifiers, we found that LBP with random classifiers results in a more substantial accuracy drop, as shown in **Fig. 3** (b) and **Fig. 4** (b). For example, in **Fig. 3** (b), at $k = 20$, applying random classifiers in LBP1 causes a 14.91% accuracy drop, whereas only 0.91% is incurred for BP.

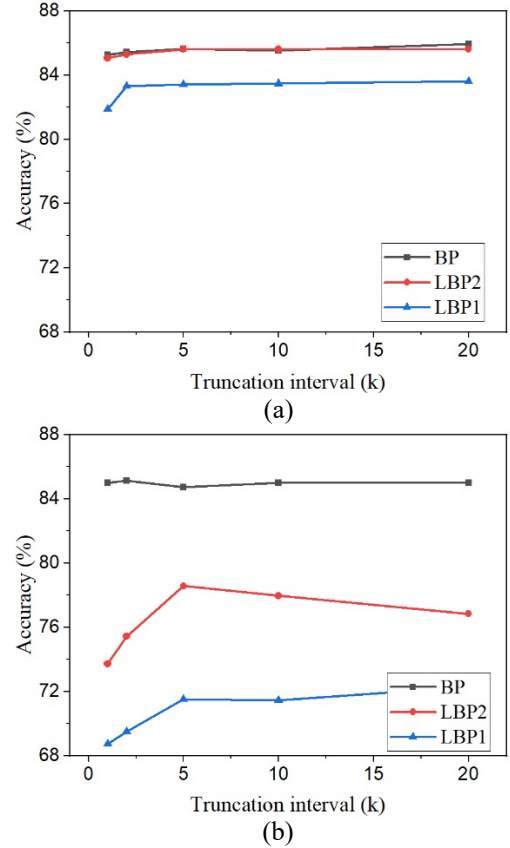

**Fig. 3** Classification accuracy of LeNet-1 on EMNIST dataset for different truncation intervals and local block size. Accuracy results are obtained with (a) trainable classifiers and (b) random classifiers. In LBPn, n indicates the number of layers in a local block.

In the simulations on DvsGesture dataset, $k$ was chosen from the set, $\{60, 30, 20, 10, 5, 1\}$, and $n$ from the set, $\{4, 2, 1\}$. In the simulations on CIFAR10-DVS dataset, $k$ was chosen from the set, $\{100, 50, 20, 10, 1\}$, and $n$ from the set, $\{8, 4, 2, 1\}$. The accuracy results of the two datasets are shown in **Fig. 5** and **Fig. 6**, respectively. Different from the frame-based datasets, the DVS-recorded datasets show different changes of accuracy with temporal truncation. The accuracy tends to increase while the truncation interval decreases and then decreases when the interval becomes small. This could be because temporal truncation with a large $k$ generally gives better convergence at the cost of a long training time, while a small $k$ can cause the network not to converge, thus resulting in bad performance [53]. This suggests that trained on the datasets containing temporal information, SNNs can benefit from temporal truncation in improving classification performance. But the optimal truncation interval varies dependent on datasets. Moreover, as shown in **Fig. 5** (a), on DvsGesture dataset, LBP leads to better accuracy than BP in the case of trainable classifiers. The same comparison can be observed in the results of CIFAR10-DVS dataset, as shown in **Fig. 6** (a) and (b). This further confirms that LBP could reduce overfitting effect because both DVS-recorded datasets contain a small number of training samples. The combining effect can be seen in **Fig. 5** (a) and **Fig. 6** (a)(b), where the improvement in LBPs is more significant than in BP. Additionally, as shown in **Fig. 5** (b) and **Fig. 6** (b), applying random classifiers in LBP1 incurs significant accuracy loss, namely, 10.99% on DvsGesture and 13.86% on CIFAR10-DVS, when $k$ is the total time step.

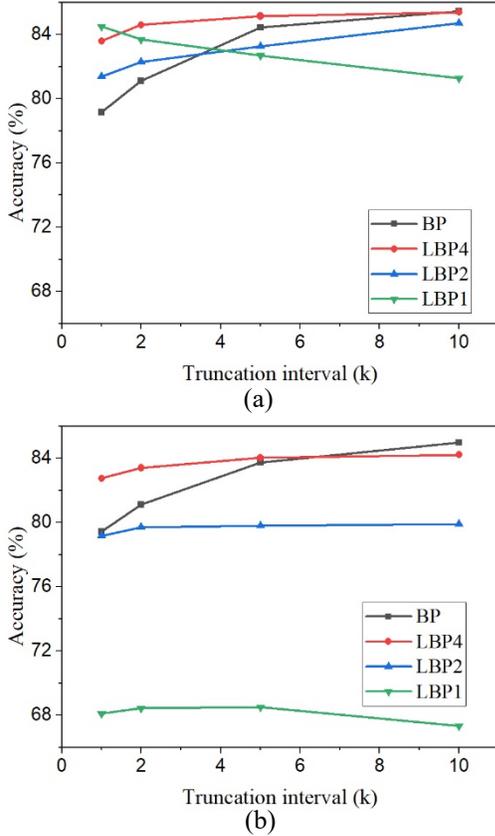

(a)

(b)

**Fig. 4** Classification accuracy of AlexNet on CIFAR10 dataset for different truncation intervals and local block size. Accuracy results are obtained with (a) trainable classifiers and (b) random classifiers.

In summary, temporal truncation affects the classification performance of SNNs differently on different types of datasets. It shows an overall negative impact on classifying frame-based datasets, whereas it brings on benefits in improving classification performance on DVS-recorded datasets with optimally chosen intervals. From all the experiments, we can see that a good choice of the truncation interval appears in the range from 2 to 10. Regarding local training, in the scenario where overfitting is not severe, such as in LeNet-1 trained on EMNIST, LBP causes inferior performance. On the other hand, LBP can alleviate overfitting effect to some extent and lead to better accuracy. The roles of temporal truncation and local training are not orthogonal. Instead, in most cases, they tend to function synergistically. In addition, the use of random classifiers results in accuracy loss, especially in LBP1.

## IV. COMPUTATIONAL COST

DNNs are typical of tens of or hundreds of layers with millions of or even billions of parameters. They are both computationally and memory intensive, making them difficult to be implemented in hardware. Training DNNs poses a great challenge in hardware design. Thus, it is essential to assess the computational cost of a training algorithm. In this section, we will analyze the computational cost of the proposed training method on different hardware platforms in terms of required GPU memory, external memory access, and number of arithmetic operations.

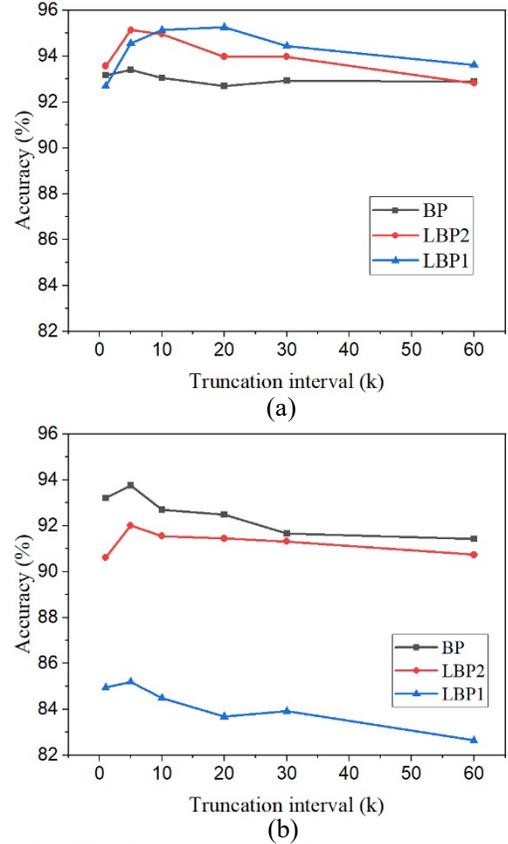

(a)

(b)

**Fig. 5** Classification accuracy of LeNet-2 on DvsGesture dataset for different truncation intervals and local block sizes. Accuracy results are obtained with (a) trainable classifiers and (b) random classifiers.

### A. GPU memory cost

Temporal truncation reduces the length of backward pass and eliminates the requirement to store the history of neural states before the current interval, which leads to memory saving. Local training avoids the necessity to build a whole backward computational graph by training SNNs block by block, which also leads to memory saving. In this section, we measure and compare the maximum GPU memory used to perform each classification task under different settings of $(k, n)$ in Pytorch. The measurement was done with the commonly-used command *max_memory_allocated* in Pytorch [35, 54].

**Fig. 7** shows the measured GPU memory cost for training LeNet with trainable classifiers on different datasets. Due to the larger network scale and longer training window, the memory cost of LeNet-2 is much higher than that of LeNet-1. Clearly, memory cost decreases linearly with decreasing truncation interval. Compared to BP, LBP2 can reduce memory cost. As shown in **Fig. 7** (a) and (b), the reduction percentage increases with decreasing interval from 20.14% to 64.73% for classifying EMNIST and from 2.06% to 32.61% for classifying

DvsGesture. Further decreasing $n$ in LBP contributes to minor change. This is mainly because the uneven distribution of layer neurons and parameters in networks causes particular layers to dominate in memory occupation. Replacing trainable classifiers with random classifiers has a negligible effect on reducing memory cost.

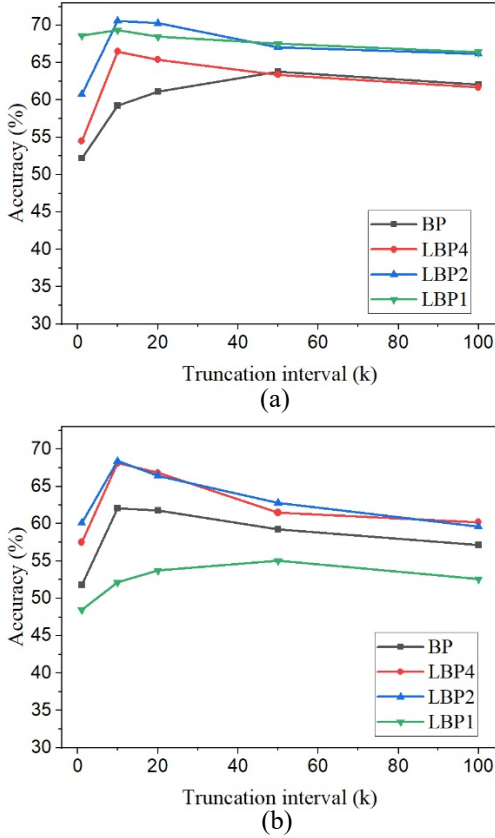

**Fig. 6** Classification accuracy of AlexNet on CIFAR10-DVS dataset for different truncation intervals and local block size. Accuracy results are obtained with (a) trainable classifier and (b) random classifier.

The same observations can be seen in the memory cost for training AlexNet on CIFAR10 and CIFAR10-DVS datasets, as shown in **Fig. 8** (a) and (b), respectively. LBP1 helps reduce the memory cost from 31.86% to 60.91% for classifying CIFAR10, and from 3.44% to 54.42% for classifying CIFAR10-DVS with the decreasing truncation interval. Using random classifiers leads to less than a 2% reduction.

In the simulation framework, the memory cost of neural networks are caused by the storage of parameters, network states, and computational graphs (CGs) [55]. The memory cost (MC) of an SNN can be determined by

$$MC = N(w) + MC(CG) + C$$
$$= N(w) + N(states) + N(inter.) + N(grads) + C \quad (8)$$

where $N(w)$, $N(states)$, $N(inter.)$, and $N(grads)$ are the number of trainable parameters, neural states, intermediate tensors allocated in computational graphs, and gradients, respectively, $C$ is a constant representing the memory consumed by CUDA workspace. In TBPTT, networks are unfolded in time, and the history of the states, intermediate tensors, and gradients are saved. Once the backward update is finished, all the history is discarded. Thus the memory cost of TBPTT with a truncation interval k is

$$MC = k\big(MC(CG)\big) + N(w) + C \quad (9)$$

The benefit of local training methods is only possible at the last time step of the truncation interval at which a partial graph of the block length is saved. When local training method is applied, the memory cost becomes

$$MC = (k-1)\big(MC(CG)\big) + MC_{Local}(CG) + N(w) + C \quad (10)$$

where $MC_{Local}(CG)$ is the maximum memory cost of local training methods determined by layer distribution in the network. $MC(CG)$ depends on network architectures and remains constant in time, which explains the linearity in the memory change. The memory gap between BP and LBP is determined by the difference between $MC(CG)$ and $MC_{Local}(CG)$, i.e. the difference between a whole graph and a partial graph. The difference is the function of network architectures and proportional to the block length, remaining unchanged with $k$.

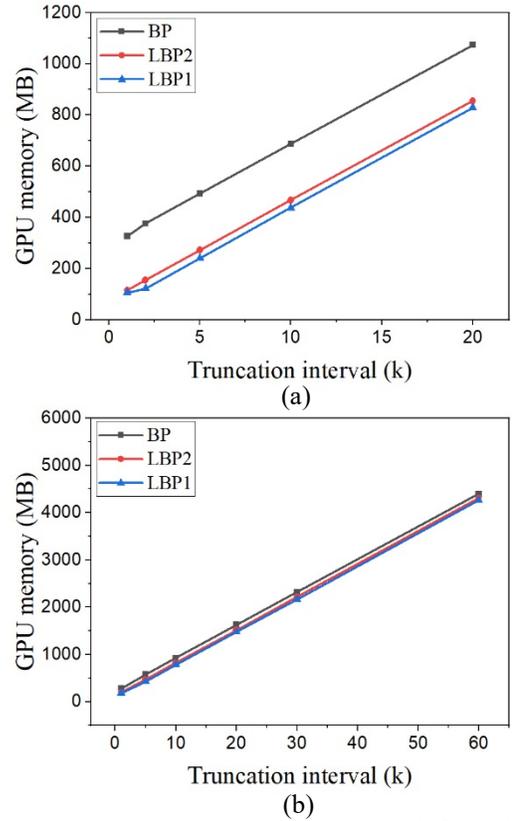

**Fig. 7** GPU memory cost measured in Pytorch for training (a) LeNet-1 on EMNIST dataset and (b) LeNet-2 on DvsGesture dataset. Both SNNs were trained with trainable classifiers. The batch size is 1024 and 32 for EMNIST dataset and DvsGesture dataset, respectively.

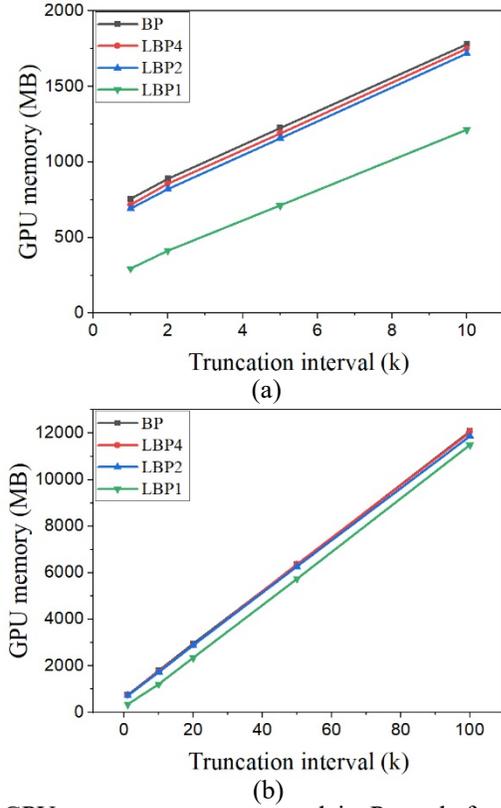

**Fig. 8** GPU memory cost measured in Pytorch for training AlexNet on (a) CIFAR10 dataset and (b) CIFAR10-DVS dataset. Trainable classifiers are used. The batch size is 128 for both cases.

*B. Memory access and arithmetic operations analysis*

The BP algorithm requires hardware to store all the neural states at each layer before performing backward updates from the top layer, as those states are needed to compute errors and gradients along the backward pass. More costly, the BPTT algorithm introduces an extra time dimension and requires the storage of the whole history of all the states at each layer. General hardware, such as CPU, GPU, and field-programmable gate array (FPGA), has limited on-chip memory capacity, which is not enough to accommodate the states and parameters of state-of-the-art networks. The intermediate states and network parameters have to be saved in external memories such as DRAM. Thus, the BPTT algorithm adds a significant memory overhead and a huge data communication burden on hardware. Frequent communication also brings on high energy consumption since memory access consumes much more energy than arithmetic operations. For example, for the 45 nm CMOS process, memory access consumes 3 orders of magnitude more energy [56]. Reducing memory access frequency can lead to significant energy and time saving. In this section, we analyze and model the memory traffic pattern and the number of arithmetic operations in training SNNs with the proposed training algorithm.

**Fig. 9** illustrates the data transfer between external memories and processing cores in both forward pass and backward pass of a local block during a truncation interval. Assume that on-chip memory has the capacity to store the parameters and batch neural states of a layer. During forward update, at time $t$, each layer has to read its weights $W^n$ and previous neural states $U^{t-1,n}$ from an external memory, and write the updated states $U^{t,n}$ back to the external memory in separate locations. We omit the transfer of spikes since they are one-bit data. The number of read and write operations in forward pass is expressed by

$$N_r^f = \sum_{n=1}^{N_l} (|W^n| + |U^n|) + (|W_c| + |U_c|),$$

$$N_w^f = \sum_{n=1}^{N_l} |U^n| + |U_c| \qquad (11)$$

where $N_l$ is the number of layers in a local block, $|W^n|$ and $|W_c|$ are the total number of weights in the layer $n$ of the main network and weights in the classifier layer, respectively, $|U^n|$ and $|U_c|$ are the total number of batch neural states in the layer $n$ and the classifier layer, respectively. During backward update, the network needs to compute the errors, $(\delta, \gamma)^{t,n}$, in each layer at each time step, and propagate the potential error backward through layers and time. At any time step in the middle of truncation interval, for example, at $t_k - 1$ in **Fig. 9**, to compute the errors at the layer $n$, the network has to read the current neural states $U^{t_k,n}$, weights from the upper layer $W^{n+1}$, and potential errors from the next time step $\gamma^{t_k,n}$. Gradients $dW^n$ also need to be read for accumulation. The updated potential errors $\gamma^{t_k-1,n}$ and gradients $dW^n$ are written back to memory. The number of read and write operations in the middle of the backward pass is expressed by

$$N_r^b = \sum_{n=1}^{N_l} (|W^n| + 2|U^n|) + \sum_{n=1}^{N_l-1} (|W^{n+1}|) + 2|W_c| + 3|U_c| \quad (12)$$

$$N_w^b = \sum_{n=1}^{N_l} (|W^n| + |U^n|) + |W_c| + |U_c| \qquad (13)$$

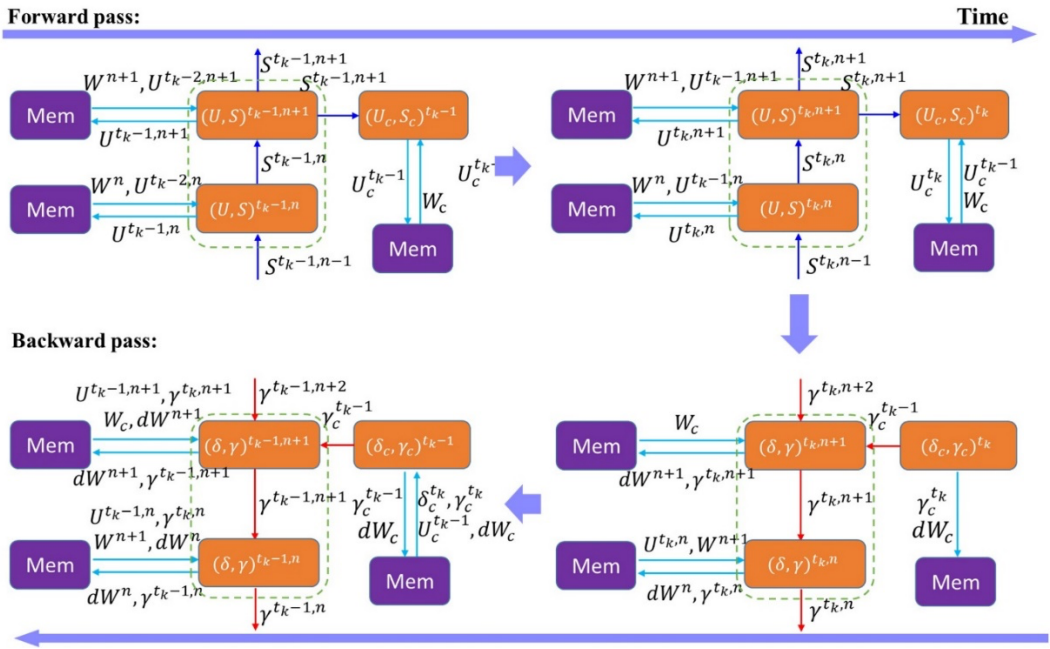

**Fig. 9** Memory traffic of forward pass and backward pass in a local learning block during a truncation interval. $(U, S)^{t,n}$ represents the neural states in the layer $n$ of the main network at time $t$. $W^n$ and $dW^n$ are the weights and gradients in the layer $n$. $(\delta, \gamma)^{t,n}$ represents the spike error and potential error in the layer $n$ at time $t$. The parameters and states of the local classifier are indicated by the subscript $c$. $t_k$ is the end of the truncation interval.

We simplify the formulation using the fact that $|dW^n| = |W^n|$ and $|\gamma^n| = |U^n|$. Clearly, in the middle of the backward pass, the local training method does not help reduce the number of memory access because the consecutive execution between the forward update and backward update is prohibited at the time step. At the end of the interval where $t = t_k$, there is no temporal component needed to compute the errors. Since the backward pass of the block can immediately start after the forward pass finishes, the current neural states of the top layer and classifier can be buffered on the chip for the backward update, eliminating the need to read them from external memory. There is also no need to read the gradients. Thus, the number of reads can be reduced to

$$N_r^b = \sum_{n=1}^{N_l-1} (|W^{n+1}| + |U^n|) + |W_c|$$

Using random weights in the classifier can further reduce memory access volume by removing all the operations on classifier weights. As argued in [32], random weights can be generated on the fly by random number generators (RNGs), avoiding the storage in external memories. It could cause either resource overhead in the case of multiple RNGs for parallel processing, or increased latency as computation needs to wait for the generation of random weights. Also, on-chip memory has to be allocated to hold them before computation finishes.

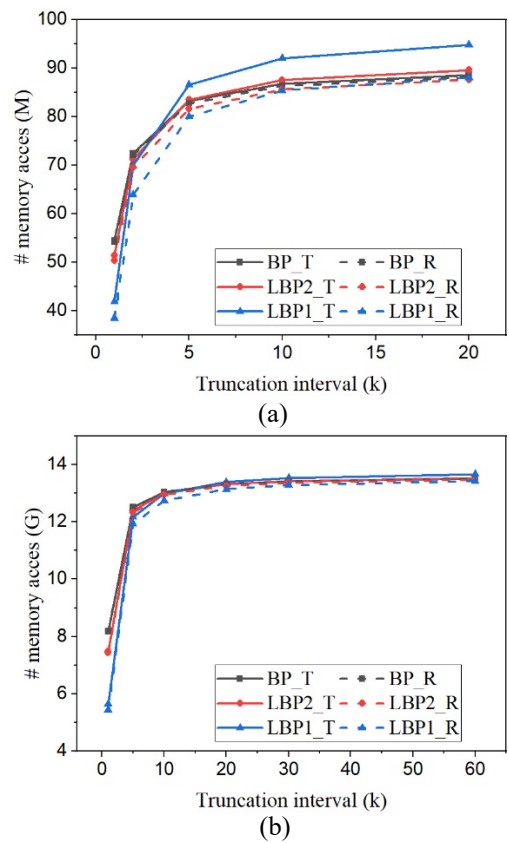

**Fig. 10** Number of memory access estimated for training (a) LeNet-1 on EMNIST dataset and (b) LeNet-2 on DvsGesture dataset for one batch iteration. Solid lines and dashed lines represent the results obtained with trainable (T) classifiers and random (R) classifiers, respectively. The batch size is 128.

SNNs replace MAC operations with multiplex-accumulation operations through event communication in the forward update. However, in the backward update, full-precision errors are the information carriers, so MAC operations are inevitable, as indicated in (4). The addition operations in the forward update are dominated by the computation of synaptic input, which is proportional to the size of weight matrices and input spike sparsity, expressed by

$$N_{add}^f = \sum_{n=1}^{N_l} \alpha^{n-1} M^n + \frac{\alpha_c |W_c||U_c|}{N_c} \qquad (14)$$

where $\alpha^{n-1}$ is the input sparsity to the layer $n$, $N_n$ and $N_c$ are the number of neurons in the layer $n$ and a classifier, respectively, and $C_o^n$ is the number of output channels in a convolutional layer $n$. $M^n$ is the total number of additions without considering sparsity, computed by $(|W^n||U^n|)/C_o^n$ for a convolutional layer and $(|W^n||U^n|)/N_n$ for a fully-connected layer. In the backward update, according to Eqs. (4),(5) and (7), the number of additions is estimated as

$$N_{add}^b = \sum_{n=1}^{N_l} (2|U^n| + N_b \alpha^{n-1} |W^n|) + 2|U_c| + N_b \alpha_c |W_c| \quad (15)$$

It is worth to note that the batch size $N_b$ is multiplied with the weight matrix size representing a computation of a batch of gradients. MAC operations only appear in (4), which are used to propagate errors backward from upper layers. The number of MACs can be expressed as

$$N_{mac}^b = \sum_{n=1}^{N_l-1} M^{n+1} + \frac{|W_c||U_c|}{N_c} \tag{16}$$

### 1) Memory access

We estimated the number of memory access, including reads and writes required in one training iteration. The batch size is kept as 128 for all the cases. **Fig. 10** (a) and (b) show the estimation for training LeNet with both types of classifiers on EMNIST dataset and DvsGesture dataset, respectively. The number of memory access decreases with the truncation interval, rapidly when the interval becomes small. With trainable classifiers, LBPs lead to more memory access when the interval is large, because of the overhead of classifier weights. When the interval is small, the advantage of LBPs becomes more significant, thus overcoming the overhead. On the contrary, the use of random classifiers avoids the overhead, making LBPs better than BP at all intervals. Specifically, on EMNIST dataset, temporal truncation can contribute to around a 55% reduction for LBP1 with either type of classifiers. LBP1 can lead to around 23% reduction with trainable classifiers and 29% reduction with random classifiers at $k = 1$ against BP, respectively.

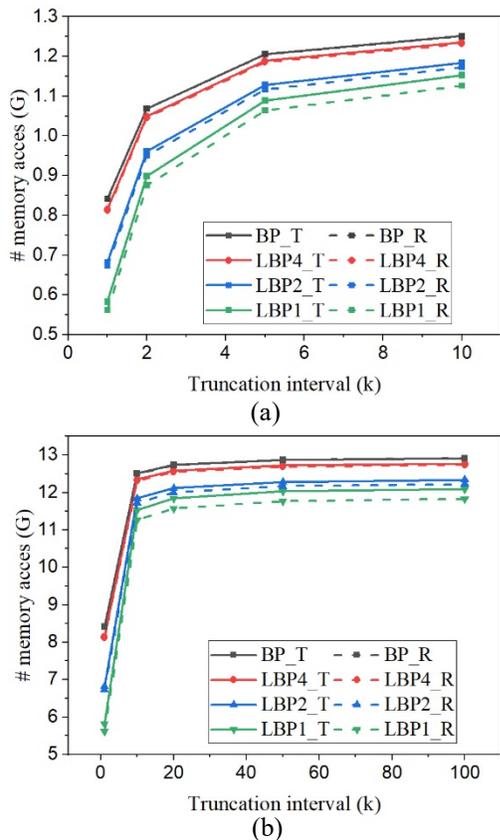

**Fig. 11** Number of memory access estimated for training AlexNet (a) on CIFAR10 dataset and (b) on CIFAR10-DVS dataset.

The estimated number of memory access for training AlexNet on CIFAR10 dataset and CIFAR10-DVS dataset is shown in **Fig. 11** (a) and (b). In AlexNet, the size of a classifier layer is much smaller compared to the network layers. The small overhead of trainable classifiers is overcome by the benefit. So LBPs lead to a reduction in memory access at all intervals. On CIFAR10, temporal truncation can lead up to around 50% reduction in LBP1. Compared against BP, LBP1 can lead up to 31% with trainable classifiers and 33% with random classifiers, respectively.

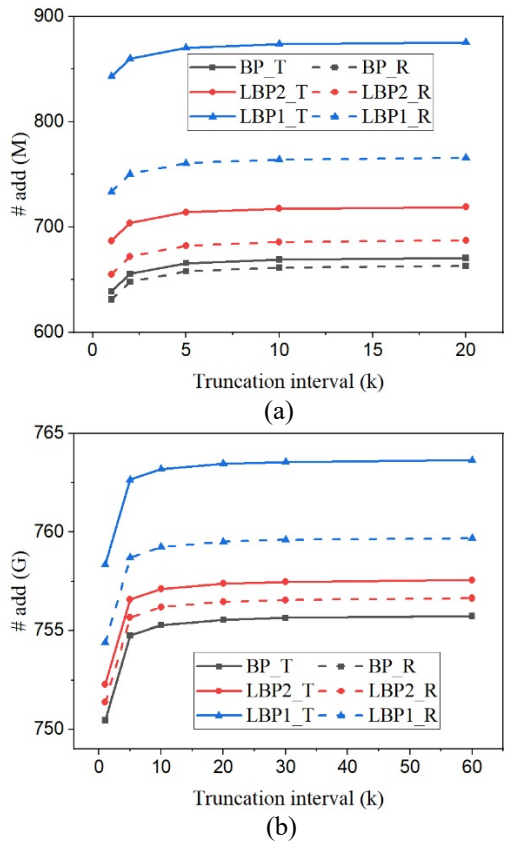

**Fig. 12** Number of additions estimated for training (a) LeNet-1 on EMNIST dataset and (b) LeNet-2 on DvsGesture dataset for one batch iteration. The batch size is 128.

### 2) Arithmetic operations

Based on the analytical model above, we estimated the number of arithmetic operations involved in one batch training iteration, including additions and MACs. **Fig. 12** (a) and (b) plot the results for LeNet trained on EMNIST dataset and DvsGesture dataset, respectively. **Fig. 13** (a) and (b) show the results for AlexNet trained on CIFAR10 dataset and CIFAR10-DVS dataset, respectively. All the results reveal the same trend of change of accuracy affected by temporal truncation and local training. Temporal truncation does not lead to a notable reduction, less than 5%/0.7% in LeNet-1/2 and 0.3% reduction in AlexNet at maximum. However, LBPs cause more additions than BP, up to 32% in LeNet-1, only 1% in LeNet-2 and 0.66% in AlexNet. This large difference is due to the proportion of classifiers in the whole network. In small networks with local classifiers, such as LeNet-1, the classifier size is comparable to the size of the main network, which causes a large overhead. The use of random classifiers can reduce the overhead to 16% in LeNet-1. Thus, random classifiers are beneficial to small networks in this regard.

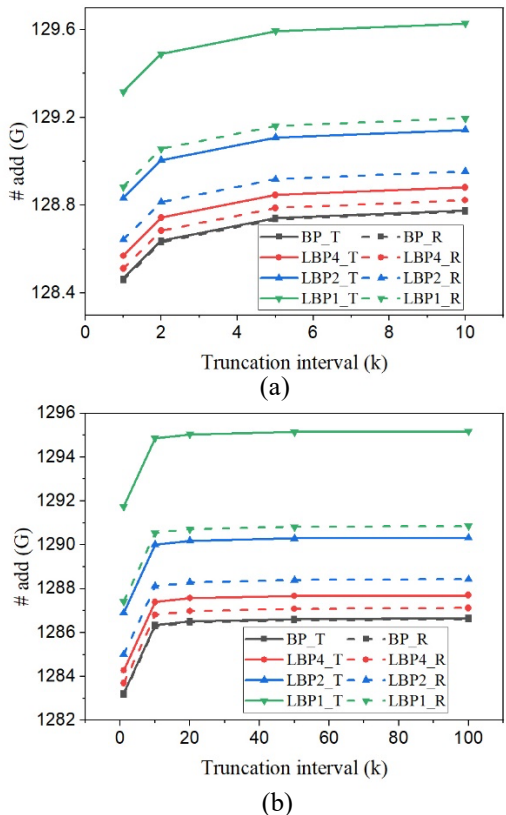

**Fig. 13** Number of additions estimated for training AlexNet (a) on CIFAR10 dataset and (b) on CIFAR10-DVS dataset.

The estimated number of MACs for different networks in one batch iteration is shown in **Fig. 14**. We normalized the values over the number of MACs required by BP in each task. The number was also averaged over the training time window. The number of MACs is independent of truncation interval and the type of classifiers but dependent on the size of networks and local training blocks. Generally, multi-channel convolutional layers consume much more MACs than linear layers. In local training methods, convolutional operations between blocks are avoided because errors are not propagated. The local error propagation is from linear classifiers, leading to very small overhead. Increasing the number of local training blocks can significantly reduce MACs. Specifically, LBP1 leads to a 72% reduction in LeNet-1 and a 99% reduction in both LeNet-2 and AlexNet. The significant reduction in MACs is one of the most attractive benefits of local training methods, as it can greatly improve the training energy efficiency of SNNs and is not affected by BPTT.

## V. CONCLUSIONS AND DISCUSSIONS

We have investigated and analyzed the impact of the design variables on classification performance and computational cost in various tasks. In this section, we will address the important design problem with regard to the optimal choice of the variables while considering many performance aspects. The role of random classifiers will be discussed. Then, we will discuss the limitations of our training method and the promising solutions.

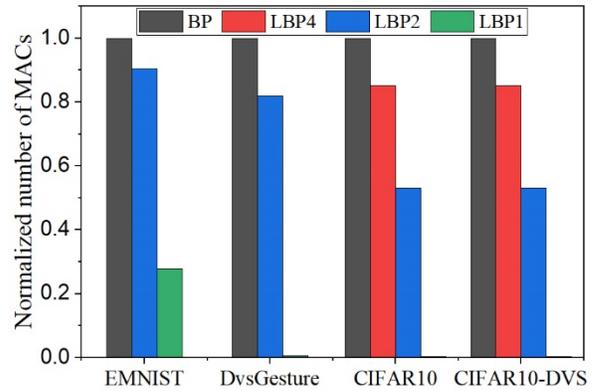

**Fig. 14** Normalized number of MACs estimated for training different networks averaged over training time windows for one batch iteration. For each task, the number of MACs was normalized over that required by BP.

### A. Brief summary

We have studied what roles the temporal truncation and local training play in affecting accuracy and computational cost including GPU memory cost, memory access, and arithmetic operations. The design space regarding the length of truncation interval and the size of local training blocks was explored. The impact of temporal truncation on accuracy depends on the type of datasets. It tends to decrease accuracy on frame-based datasets, while improves accuracy on DVS-recorded datasets with properly chosen intervals. Local training harms the classification performance when the size of network fits well with datasets, whereas it leads to improvement in the accuracy of the networks when overfitting is severe. In most cases, temporal truncation functions synergistically with local training. The combined effect helps slow down the decrease of accuracy and even improve accuracy in many cases. Both methods can contribute to a substantial reduction in GPU memory. Temporal truncation reduces memory access volume and has a negligible effect in lessening computational operations. Local training causes notable overhead in memory access and additions in small networks. However, it brings down the number of MACs remarkably.

### B. How to determine the design variables?

It remains challenging in how to choose the degrees of temporal truncation and spatial locality, i.e., the values of $(k, n)$. The choice depends on classification tasks and also the trade-off between classification performance and computational cost. For good classification performance, local training method could be promising with the block length larger than 1 and a good choice of k lies in the range from 2 to 10. For low computational cost, the best choice is undoubtedly the layer-wise local training with the truncation interval of 1. To provide a guidance for selecting $(k, n)$, we define a figure of merit (FoM) considering both accuracy and computational cost equally as below

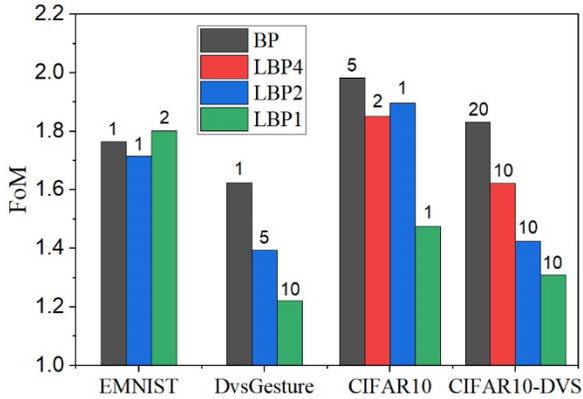

**Fig. 15** FoMs for different local training methods on different datasets. For each local training method, the smallest FoM is selected. On top of each bar, the value represents the truncation length.

$$FoM = AL + 0.25 * (MC + \#MA + \#ADD + \#MAC) \quad (17)$$

where $AL$ is the accuracy loss, $MC$ is the GPU memory cost, $\#MA$, $\#ADD$, and $\#MAC$ are the number of memory access, additions, and MAC, respectively. All the terms are normalized against the BPTT method. From the definition, a small FoM is desirable. **Fig. 15** displays the comparison among different local training methods across all the datasets under the defined FoM. For each local training method, the smallest FoM is selected, and the corresponding value of $k$ is shown on top of each bar. From the comparison, the layer-wise local training method (LBP1) shows the best FoM on all the datasets except for EMNIST. In most cases, the best $k$ lies in the range from 1 to 10. Specifically, the best values of $(k, n)$ are $(1, 2)$, $(10, 1)$, $(1, 1)$, and $(10, 1)$ on EMNIST, DvsGesture, CIFAR10, and CIFAR10-DVS, respectively. It is worth noting that the proposed FoM considers the equal contribution from accuracy and computational cost and different definitions can be proposed to determine the design choice under practical application constraints.

In **Fig. 16** (a) (b), we summarized the accuracy drop and computational cost reduction of the best training design according to the proposed FoM. The BPTT method is taken as the baseline for calculating the accuracy drop and computational cost reduction. On EMNIST and CIFAR10, the accuracy drop is within 1%, whereas the accuracy can be improved by up to 7.26% in the other cases. On the other hand, the proposed training method leads to >80% reduction in GPU memory cost and >99% reduction in the number of MACs in most cases. On two datasets, the number of memory access is also considerably reduced by >40%. A negligible overhead in additions can be observed. Therefore, the proposed training method has been demonstrated to retain good classification performance or even improve it while achieving significant reduction in computational cost.

### C. The role of random classifiers

The use of random classifiers in LBPs was proposed for its potential contribution in reducing computational cost in both ANNs and SNNs [32, 36]. No comparisons were made between random classifiers and trainable classifiers. Our work reveals a detailed comparison between them. Random classifiers cause worse accuracy when used with LBPs. Especially, the loss becomes more severe in LBP1. They are beneficial in reducing memory access and additions in small networks, but make negligible contributions to reduction in GPU memory, memory access, and additions in large networks. They have no effect on MACs. More specifically, in AlexNet trained on CIFAR10-DVS with the same $(k, n)$, random classifiers can cause 20.16% accuracy drop with negligible improvement in computational cost compared to trainable classifiers. Therefore, our study shows that trainable classifiers have more considerable merit than random classifiers.

### D. Limitations and future perspectives

The temporal truncation interval and local block length are two hyper-parameters that requires to be optimized in our proposed method. In our study, we have adopted a grid search to determine the optimal values. The results reflect that the optimal values vary from task to task and network to network, which poses a limitation on the applicability of the proposed method. A good choice of $(k, n)$ has been discussed above under the proposed FoM. Although the optimal performance may not be achieved on all the tasks, it can still deliver promising improvement. Hyperparameter optimization has remained a challenge in deep learning. Most commonly in

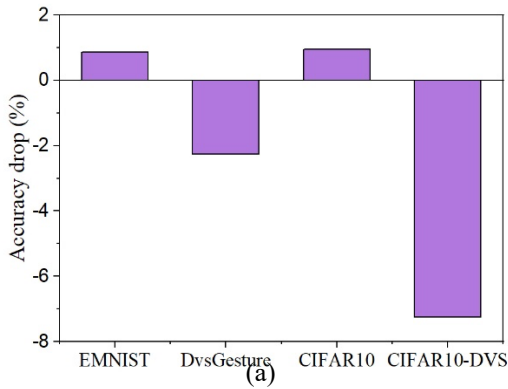

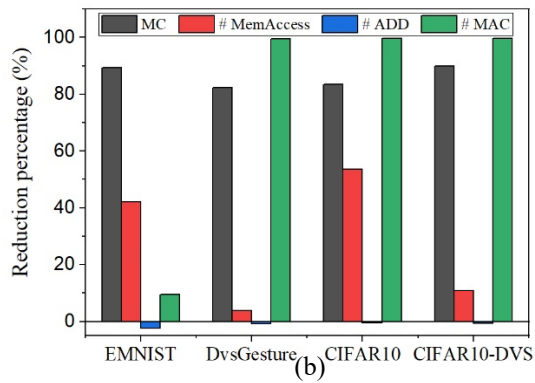

**Fig. 16** (a) Accuracy drop and (b) computational cost reduction of the selected method by the proposed FoM on all the datasets. The BPTT method is taken as the baseline for calculating the accuracy drop and computational cost reduction.

literatures, hyperparameters are chosen based on rules of thumb summarized in practice involving manual tuning. There exists many optimization algorithms. Classic approaches, such as random search [57], Bayesian model [58], and evolutionary algorithms [59], are generally time consuming and may not converge. In recent years, gradient-descent based optimization methods have made it possible to directly optimize hyperparameters in the training loop, such as bilevel optimization [60]. Thus, such optimization method could be a promising addition to our proposed training method to automate hyperparameter search for achieving optimal performance in various tasks.

TBPTT can be implemented in different ways. Instead of going through all the time steps in the backward pass during a truncation interval, the backward update can stop in the middle. In other words, the backward update can have a shorter time pass than the forward update. Cutting short the backward pass can furthermore reduce computational cost. Although in our work local training has been shown to improve classification performance in some cases, the intrinsic downside of local training method still remains and could considerably harm the performance of large-scale networks such as ResNets for more complex tasks such as ImageNet. As pointed out by Wang *et al.*, local training is short-sighted and suffers from essential information loss while progressing along with the network [35]. Many solutions were proposed to alleviate this issue. Wang *et al.* proposed an alternative loss function considering information preservation [35]. Nokland *et al.* applied an auxiliary loss function to create another backward pass for information flow [34]. These proposals provide opportunities for further improvement in classification performance in our proposed training method.

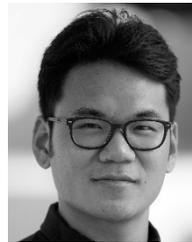

**Wenzhe Guo** received his B.S. majored in integrated circuits design from University of Electronic Science and Technology of China (UESTC), Chengdu, China, in 2017. He obtained M. Sc. degree in Electrical Engineering department in King Abdullah University of Science and Technology (KAUST), Thuwal, Saudi Arabia, in 2018. Currently, he is a Ph.D. student in KAUST, working on the design of neuromorphic hardware systems for different applications. His research interests include digital IC design, neuromorphic computing, event-based computation, and computer vision.

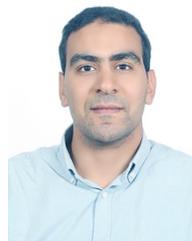

**Mohammed E. Fouda** received the B.Sc. degree (Hons.) in Electronics and Communications Engineering and the M.Sc. degree in engineering mathematics from the Faculty of Engineering, Cairo University, Cairo, Egypt, in 2011 and 2014, respectively. Fouda received his Ph.D. degree from the University of California-Irvine, USA in 2020. Currently, He works as an assistant researcher at University of California, Irvine. Fouda published more than 120 peer-reviewed Journal and conference papers, 1 Springer book and 3 book chapters. His H-index is 22 with more than 1800 citations. His research interests include analog AI hardware, neuromorphic circuits and systems, brain-inspired computing, memristive circuit theory, fractional circuits and systems and analog circuits. He serves as a peer-reviewer for many prestigious journals and conferences. He also serves as an associate editor International Journal of Circuit theory and applications in addition to a technical program committee member in many conferences. He was the recipient of the best paper award in ICM years 2013 and 2020 and the Broadcom foundation fellowship for 2016-2017.

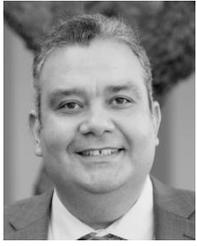

Ahmed M. Eltawil is a Professor of Electrical and Computer Engineering at King Abdullah University of Science and Technology (KAUST) where he joined the Computer, Electrical and Mathematical Science and Engineering Division (CEMSE) in 2019. Prior to that he was with the Electrical Engineering and Computer Science Department at the University of California, Irvine (UCI) since 2005. At KAUST, he is the founder and director of the Communication and Computing Systems Laboratory (CCSL). His current research interests are in the general area of smart and connected systems with an emphasis on mobile systems. He received the Doctorate degree from the University of California, Los Angeles, in 2003 and the M.Sc. and B.Sc. degrees (with honors) from Cairo University, Giza, Egypt, in 1999 and 1997, respectively. Dr. Eltawil has been on the technical program committees and steering committees for numerous workshops, symposia, and conferences in the areas of low power computing and wireless communication system design. He received several awards, including the NSF CAREER grant supporting his research in low power computing and communication systems. He is a senior member of the IEEE and a senior member of the National Academy of Inventors, USA. In 2021, he was selected as "Innovator of the Year" by the Henry Samueli School of Engineering at the University of California, Irvine where he received US congressional recognition for his contributions.

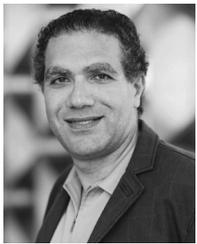

Khaled Nabil Salama (SM'10) received the B.S. degree from the Department Electronics and Communications, Cairo University, Cairo, Egypt, in 1997, and the M.S. and Ph.D. degrees from the Department of Electrical Engineering, Stanford University, Stanford, CA, USA, in 2000 and 2005, respectively. He was an Assistant Professor at Rensselaer Polytechnic Institute, Troy, NY, USA, from 2005 to 2009. He joined the King Abdullah University of Science and Technology (KAUST), Thuwal, Saudi Arabia, in 2009, where he is currently a Professor, and was the Founding Program Chair until August 2011. He is the Director of the sensors initiative, a consortium of nine universities (KAUST, MIT Cambridge, UCLA, GATECH, Brown University, Georgia Tech, TU Delft, Swansea University, the University of Regensburg, and the Australian Institute of Marine Science (AIMS)). His work on CMOS sensors for molecular detection has been funded by the National Institutes of Health (NIH) and the defense advanced research projects agency (DARPA), awarded the Stanford–Berkeley Innovators Challenge Award in biological sciences and was acquired by Illumina Inc. He is the author of 250 articles and 14 U.S. patents on low-power mixed-signal circuits for intelligent fully integrated sensors and neuromorphic circuits using memristor devices.